# Multidimensional precipitation index prediction based on CNN-LSTM hybrid framework

Yuchen Wang [1,4], Pengfei Jia*[1,5], Zhitao Shu[2,6], Keyan Liu[3,7], Abdul Rashid Mohamed Shariff[1,8]

[1] Department of Biological and Agricultural Engineering, Universiti Putra Malaysia, Malaysia
[2] Vanderbilt University, Nashville, TN, USA
[3] Arizona State University, Houston, TX, USA

[4] wyuchen080@gmail.com

[5] jiapengfei97@gmail.com

[6] zhitao.shu@vanderbilt.edu

[7] kliu121@asu.edu

[8] gs69687@student.upm.edu.my

**Abstracts.** With the intensification of global climate change, accurate prediction of weather indicators is of great significance in disaster prevention and mitigation, agricultural production, and transportation. Precipitation, as one of the key meteorological indicators, plays a crucial role in water resource management, agricultural production, and urban flood control. This study proposes a multidimensional precipitation index prediction model based on a CNN-LSTM hybrid framework, aiming to improve the accuracy of precipitation forecasts. The dataset is sourced from Pune, Maharashtra, India, covering monthly mean precipitation data from 1972 to 2002. This dataset includes nearly 31 years (1972-2002) of monthly average precipitation, reflecting the long-term fluctuations and seasonal variations of precipitation in the region. By analyzing these time series data, the CNN-LSTM model effectively captures local features and long-term dependencies. Experimental results show that the model achieves a root mean square error (RMSE) of 6.752, which demonstrates a significant advantage over traditional time series prediction methods in terms of prediction accuracy and generalization ability. Furthermore, this study provides new research ideas for precipitation prediction. However, the model requires high computational resources when dealing with large-scale datasets, and its predictive ability for multidimensional precipitation data still needs improvement. Future research could extend the model to support and predict multidimensional precipitation data, thereby promoting the development of more accurate and efficient meteorological prediction technologies.

# 1. Introduction

As global climate change becomes increasingly severe, frequent meteorological disasters have had a significant negative impact on society and the environment. The analysis and prediction of meteorological data have a wide range of applications in disaster prevention and mitigation, agricultural production, and transportation. Among various meteorological factors, precipitation is one of the most critical indicators, and its prediction is crucial for water resource management, agricultural production, urban flood control, and other fields. However, due to the high complexity and uncertainty of precipitation data, traditional precipitation prediction methods face numerous challenges, such as large computational requirements, limited accuracy, and slow response times. Traditional statistical methods and physical equations are increasingly unable to meet the current demands for changing precipitation prediction.

Over the past few decades, precipitation prediction has been a focal point of research in the meteorological field. Precipitation is influenced by multiple factors, including atmospheric circulation, temperature, and humidity, and exhibits clear seasonal variations and short-term fluctuations. This makes precipitation forecasting particularly difficult during extreme weather events. Traditional methods for predicting precipitation mainly rely on regression analysis, time series models, and numerical weather prediction (NWP) techniques [1]. However, these methods often perform poorly when faced with complex spatiotemporal relationships and multivariable data, especially in small-scale and local precipitation forecasts.

In recent years, the rapid development of deep learning techniques has opened up new possibilities for precipitation prediction. Compared with traditional meteorological prediction methods, deep learning models are better at capturing the complex nonlinear relationships and spatiotemporal patterns in precipitation data. By transforming precipitation prediction into a spatiotemporal sequence prediction problem, deep learning methods can significantly improve prediction accuracy and efficiency. Specifically, in precipitation time series forecasting, deep learning models such as Convolutional Neural Networks (CNN) and Long Short-Term Memory networks (LSTM) can automatically learn the patterns and regularities in precipitation data from large amounts of historical data, thus improving prediction accuracy.

This study proposes a multidimensional precipitation index prediction model based on the hybrid CNN-LSTM framework, aiming to enhance the accuracy and reliability of precipitation forecasting. The model combines the advantages of CNN in feature extraction and LSTM in modeling long-term dependencies. CNN is effective at extracting temporal features of precipitation data, while LSTM captures long-term dependencies within the data. The combination of the two enables the model to better capture the complex dynamic changes in precipitation data.

The dataset used in this study is from Pune, Maharashtra, India, and includes monthly average precipitation data from 1972 to 2002. This dataset spans nearly 31 years (1972-2002) and provides monthly averages of precipitation, reflecting long-term fluctuations and seasonal variations in precipitation over the region. By analyzing this precipitation time series data, the proposed CNN-LSTM hybrid framework effectively captures both temporal features and long-term dependencies in precipitation, offering a new approach to multidimensional precipitation forecasting.

# 2. Related work

Time series forecasting is crucial in weather forecasting, financial - market analysis, and traffic flow prediction. In meteorology, precipitation forecasting is vital for water resource management, farming, and urban flood prevention. However, due to the complexity and nonlinearity of time series data, especially during extreme weather and with multidimensional meteorological data, traditional forecasting methods face challenges in accuracy and generalization.

In recent years, deep learning technologies have made remarkable progress in precipitation forecasting.Chen G et al [2]. developed a 3D CNN for short-term precipitation prediction over the contiguous United States, using 39 years of meteorological and daily precipitation data. Their model outperforms state - of - the - art weather models in predicting daily total precipitation, with superior performance up to 5 day forecasts. Combining the model's predictions with traditional weather forecasts significantly enhances forecast accuracy, especially for heavy precipitation events. The network's millisecond scale inference time also enables large ensemble predictions for further accuracy improvement, demonstrating deep learning's potential in weather prediction.

Jin W et al [3]. developed a deep learning based model for seasonal precipitation prediction over China. The model was pre-trained on GCM hindcasts and fine-tuned with ERA5 and observational data. It showed improved PCCs of 0.71, outperforming raw GCM outputs by 0.10 - 0.13. Used at China's National Climate Center for two years, it provided effective guidance for summer precipitation prediction.

Manna T et al [4]. integrated Rough Set on Fuzzy Approximation Space (RSFAS) with Deep Learning (DL) to predict precipitation levels in India's southern coastal areas. RSFAS addresses data uncertainty, while DL handles classification and prediction. Their model, evaluated via various metrics and optimizers, outperformed existing DL and ML techniques in prediction accuracy and error reduction. It helps monitor climatic conditions during extreme events, offering early warnings of heavy rainfall to coastal residents.

Rojas-Campos A et al [5]. proposed deep learning models to generate precipitation maps using NWP's full variable set. They trained and evaluated five models, finding that these models significantly improved precipitation predictions, doubled the resolution, and corrected biases. However, some models introduced new biases. The CGAN offered the highest-quality forecasts, indicating future research directions.

Huang et al. [6] proposed a wavelet-NAR neural network-based time series prediction model for meteorological elements, which combines the advantages of wavelet analysis and nonlinear autoregressive neural network, extracts the features and removes the noise through multi-scale decomposition, effectively captures the nonlinear dynamic relationship, and the prediction results are better than those of other models, which can provide a scientific basis for the development of weather derivatives. C.D. Tian et al. [7] proposed a combined kernel function least squares support vector machine (LSSVM) prediction model, which uses a combination of polynomial function and radial basis function to construct the kernel function, and uses an improved genetic algorithm to optimise the parameters, which has the advantages of both global and local kernel function and improves the model's generalisation ability, making it superior to the traditional LSSVM model with a single kernel function and other common prediction models. Jianxin Sun et al. [8] proposed a hybrid model for dynamic correlation time series, combining ARIMA model, multiple codecs and CNN, and weighted and summed by softmax function to get the final prediction results.

## 3. Data Introduction

*3.1 Data source*
The dataset comes from the Pune region in Maharashtra, India, covering monthly mean precipitation data from 1972 to 2002, with a total of 31 years of time series data. The main feature of the data is the monthly mean precipitation, which reflects the long-term fluctuations and seasonal variation trends of precipitation in the region. Through this data, we can study the precipitation patterns in different seasons and the long-term precipitation trends, providing fundamental data for training weather prediction models.

The precipitation data in the dataset exhibits significant seasonal characteristics, effectively capturing the fluctuations in precipitation and the occurrence of extreme weather events in the region. However, due to observational or recording issues, there are some missing values in the data. During the data processing, missing precipitation values are marked as "-99". To ensure data quality, all missing values have been removed during the data cleaning process, ensuring the integrity and reliability of the dataset.

*3.2 Descriptive statistical analysis*
Figure 1 visually represents the distribution of monthly precipitation across all years in the dataset, covering the period from 1972 to 2002. The x-axis shows the months of the year, while the y-axis indicates precipitation in millimeters. From the plot, we observe a distinct seasonal pattern in precipitation, with sharp peaks between June and September, indicating a monsoon season typical for the Pune region. These months consistently show higher precipitation, although the amounts vary from year to year. In contrast, the months from January to April show relatively low and stable precipitation, representing the dry season. The final months of the year (October to December) also experience a decrease in precipitation, marking the end of the rainy period. The variability in precipitation during the wet season highlights the influence of seasonal climatic factors, which are essential for improving precipitation prediction models by capturing these seasonal and yearly fluctuations.

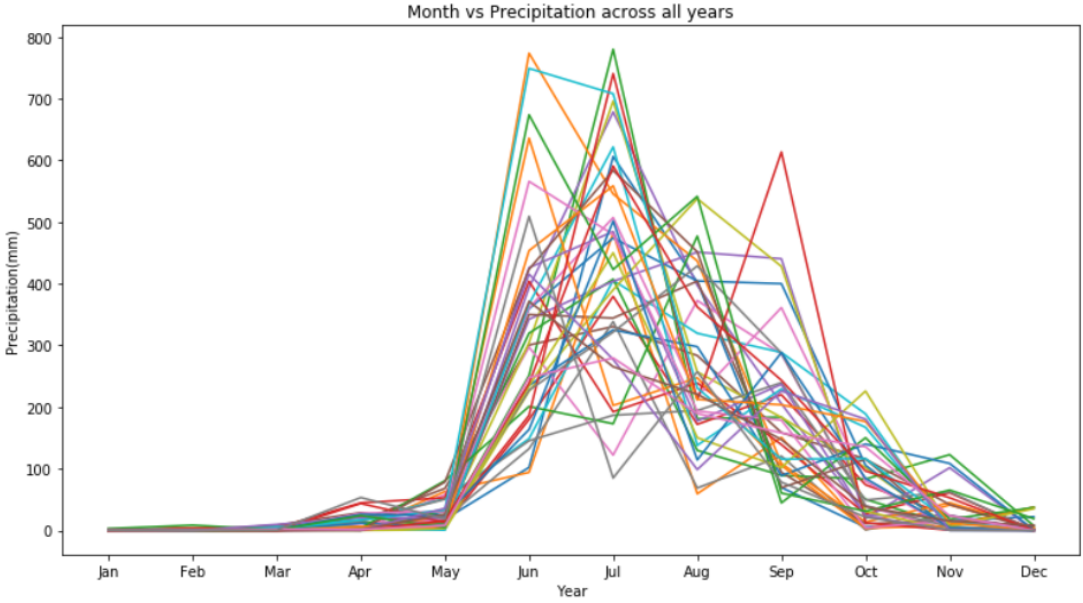

**Figure 1.** Yearly Precipitation Trends Across Different Months

Figure 2 presents a boxplot showing the distribution of monthly precipitation across all years in the dataset. The x-axis represents the months from January to December, while the y-axis indicates precipitation in millimeters. From the plot, it is clear that the rainfall in November, December, January, February, March, and April is very low, with January and February experiencing almost zero precipitation, reflecting the dry season. In contrast, June, July, and August show significantly higher precipitation, with these months experiencing much higher rainfall compared to the rest of the year, indicating the influence of the monsoon season. The range of precipitation during these months is also broader, suggesting variability in rainfall even during the monsoon period, influenced by various climatic factors each year. This boxplot highlights the seasonal effect on precipitation, emphasizing the clear distinction between the wet and dry seasons, which is crucial for improving precipitation forecasting models.

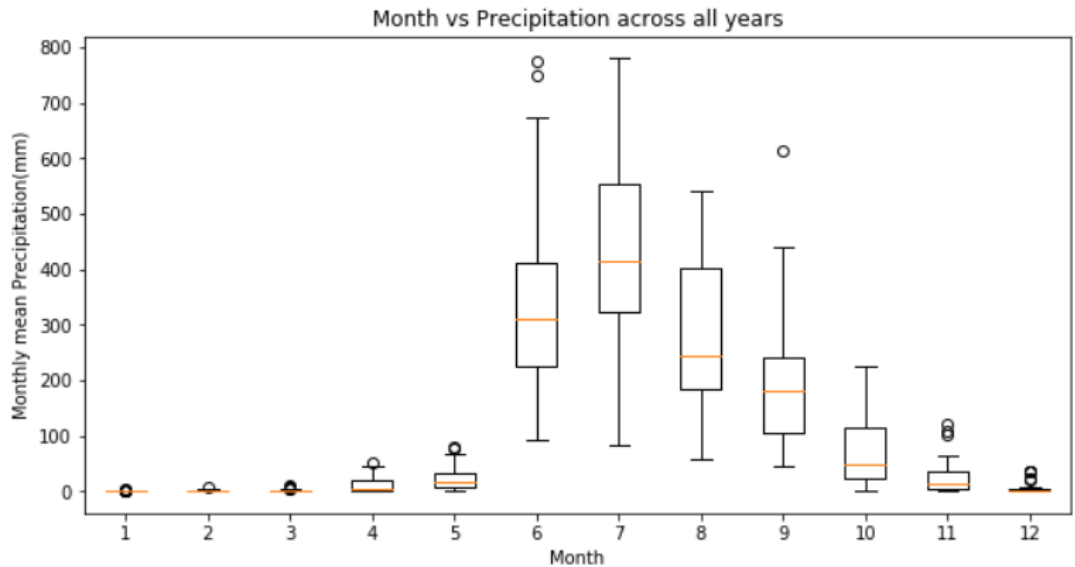

**Figure 2.** Boxplot of Monthly Precipitation Distribution Across All Years in the Pune Region

**4. CNN-LSTM Fusion Framework**

*4.1. CNN*
Convolutional Neural Networks (CNN) are a type of feedforward neural network, particularly effective in processing data with grid-like structures, such as image data. The core advantages of CNN lie in its local perception and weight sharing. In the task of precipitation prediction, CNN automatically learns the local features of precipitation data through its convolutional layers, such as seasonal fluctuations and patterns of extreme precipitation events. By applying sliding convolution operations with different kernels on time-series data, CNN can extract various representative local patterns, which are crucial for recognizing the changing trends in precipitation. For example, specific patterns of precipitation distribution in certain seasons or months can serve as unique identifiers of particular precipitation types, and CNN can effectively capture these local features [9].

In this study, CNN is used to extract time-series features from precipitation data, leveraging the convolutional layers to identify patterns at different time scales. Specifically, CNN is able to recognize short-term fluctuations and seasonal characteristics in precipitation, which are essential for modeling long-term precipitation trends. Additionally, the pooling layer further reduces the dimensionality of the feature maps output by the convolutional layers, while retaining key features. This reduction in data size and computation enhances the training efficiency and generalization ability of the model. By incorporating CNN into the CNN-LSTM hybrid framework, this study effectively extracts local features from precipitation data and combines them with the LSTM module to capture long-term dependencies in precipitation.

*4.2. LSTM*
As a special type of recurrent neural network (RNN), LSTM excels in handling time - series data, making it highly effective for precipitation prediction. Its gate mechanisms help address the gradient vanishing and exploding problems in traditional RNNs, enabling it to capture long - term dependencies in sequences. This is particularly useful for precipitation data, which shows clear seasonality and long-term trends [10].

In the CNN-LSTM hybrid framework proposed in this study, LSTM processes time-series data. CNN first extracts spatial features from the input, and the features are then fed into the LSTM network. LSTM's cell states and gates allow it to selectively retain or discard information, effectively capturing temporal dependencies like seasonality and trends in precipitation data.

During training, LSTM adjusts its weights and biases via backpropagation to minimize prediction errors. In precipitation prediction, LSTM learns the relationships between different time steps, enabling accurate future precipitation forecasting. Experiments show that LSTM can capture complex patterns in time-series data, enhancing the accuracy and reliability of precipitation predictions.

*4.3.CNN-LSTM Fusion Framework*

Model construction is an important part of the research investigation of weather indicator prediction models. In this paper, we constructed hybrid model Conv1D and LSTM. The main role is used to extract the local features of the time series data. The presence of a convolutional layer and pooling layer enables the model to automatically learn the spatial and in the original dataCNN temporal features, and to perform the dimensionality reduction and compression of the feature maps based on capturing the local features. In the model, the number of convolution kernels is 32, the kernel size is 5, the convolution step size is 1, and the Relu activation function used is to introduce nonlinearity, padding="causal".

LSTM, on the other hand, is primarily used to model long-term dependencies in time-series data by introducing three key gating structures; the forgetting gate determines the amount of information that needs to be discarded based on the current input and previous memory states. The input gate then determines what new information is worth being written to the state. Finally, the output gate combines the updated state and the current input to generate a prediction, which to some extent solves traditional gradient vanishing and gradient explosion problems when dealing with long sequential data RNNs. In this study, two layers of LSTM are by setting stacked the parameter values of units to LSTM 64 and 60 and return_sequences=True in the two models respectively. In addition, are also two fully connected layers used, the first containing 30 neurons and the second containing 10 neurons, and the activation functions are both ReLU, so that the last fully connected layer outputs only 1 value, which represents the predicted precipitation and the Lambda layer scales the output value to the original precipitation range. The period training epoch is set to 50, and the results are shown in Table 1.

**Table 1.** Model Training Parameters

| mould | loss function | optimizer | learning rate strategy | batch size | window size | training cycle |
|---|---|---|---|---|---|---|
| CNN-LSTM model | Huber | Adam | Initial 1e-8; increases 10 times every 20 epochs. | 256 | 64 | 50 |

**5.Experiments**

Figure 3 presents the performance of the CNN-LSTM model in predicting monthly precipitation on the test dataset from the Pune region. The graph plots actual future precipitation values against the model's forecasts. The lines for actual and predicted precipitation follow similar trends, especially in capturing the peaks and troughs of rainfall across the years. This shows the model's ability to recognize seasonal patterns and long-term trends in precipitation data. However, there are discrepancies in certain years where the forecast deviates from the actual values, indicating potential limitations in handling specific data fluctuations or extreme events. Overall, the model demonstrates reasonable accuracy in precipitation prediction, effectively capturing the general pattern of rainfall variation and providing a reliable reference for weather forecasting tasks.

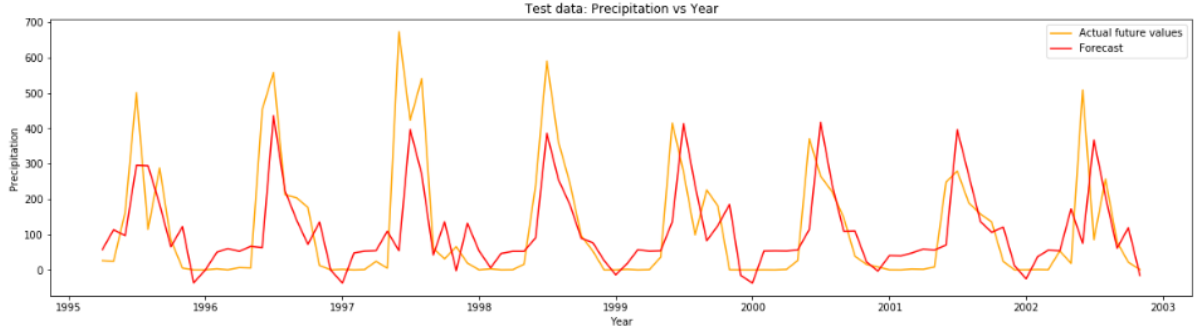

**Figure 3.** The predicted precipitation value of CNN-LSTM

The specific evaluation indicators for the CNN-LSTM model include mean square error and root mean square error. The formula is as follows:

$$\text{MSE} = \frac{1}{n}\sum_{i=1}^{n}\left(y_i - \hat{y}_i\right)^2 \quad (1)$$

$$\text{RMSE} = \sqrt{\text{MSE}} \quad (2)$$

These metrics measured the prediction error of the CNN-LSTM model on the test dataset of the precipitation data. The calculated Mean Squared Error (MSE) was approximately 45.59, and the Root Mean Squared Error (RMSE) was 6.752. Given that the monthly mean precipitation data in this experiment ranged from 0 to around 700 mm, the MSE and RMSE values indicate that the model's predictions were relatively close to the actual values. This suggests that the CNN-LSTM model has a decent performance in predicting precipitation, effectively capturing the general trend and seasonal variations of rainfall in the Pune region. However, there is still room for improvement, especially in reducing the prediction errors for years with significant precipitation fluctuations or extreme weather events.

## 6.Conclusions

This study proposes a multidimensional precipitation index prediction model based on the CNN-LSTM hybrid framework and conducts a systematic experimental investigation. By analyzing monthly precipitation data from Pune, Maharashtra, India, from 1972 to 2002, the experimental results show that the CNN-LSTM model performs well in precipitation prediction tasks. Specifically, the model's performance on the test dataset indicates that the Mean Squared Error (MSE) is approximately 45.59, and the Root Mean Squared Error (RMSE) is 6.752. Considering that the monthly mean precipitation data in the experiment range from 0 to around 700 mm, these MSE and RMSE values suggest that the CNN-LSTM model's predictions are relatively close to the actual values and are able to capture the overall trend and seasonal variations of precipitation in the Pune region effectively. However, for years with significant precipitation fluctuations or extreme weather events, the model's prediction errors still need to be further reduced.

Although the CNN-LSTM model can effectively capture the seasonal features and long-term trends of precipitation, its predictive ability remains limited when facing extreme weather events, especially in years with significant changes in precipitation. Future research can further optimize the model structure, such as incorporating additional data features or more complex network architectures, to improve prediction accuracy for extreme weather events. Moreover, given the high computational demands of large-scale dataset processing, future studies could explore lighter versions of the model to enhance its efficiency and applicability in practical scenarios.